# Performance of Recent Large Language Models for a Low-Resourced Language

Ravindu Jayakody
Department of Computer Science & Engineering
University of Moratuwa
Sri Lanka
mihirangajadr.23@uom.lk

Gihan Dias
Department of Computer Science & Engineering
University of Moratuwa
Sri Lanka
gihan@uom.lk

*Abstract*—Large Language Models (LLMs) have shown significant advances in the past year. In addition to new versions of GPT and Llama, several other LLMs have been introduced recently. Some of these are open models – available for download and modification.

Although multilingual large language models have been available for some time, their performance on low-resourced languages such as Sinhala has been poor.

We evaluated four recent LLMs on their performance directly in the Sinhala language, and by translation to and from English. We also evaluated their fine-tunability with a small amount of fine-tuning data.

Claude and GPT 4o perform well out-of-the-box and do significantly better than previous versions. Llama and Mistral perform poorly but show some promise of improvement with fine tuning.

## I. Introduction

Natural Language Processing (NLP) has seen significant advancements in recent years, largely driven by the development of large language models (LLMs) such as GPT, BERT, and LLAMA. These models, trained on massive amounts of text data, have demonstrated remarkable capabilities in various NLP tasks, including text generation, summarization, translation, and question answering. However, these models have been primarily trained on high-resource languages such as English, raising concerns about their performance on low-resourced languages with limited availability of data.

Sinhala, the official language of Sri Lanka, is one such low-resourced language that has received less attention in the NLP community. Despite being spoken by over 16 million people, the availability of Sinhala language resources – such as corpora, datasets, and pre-trained models – is limited. This scarcity of resources poses significant challenges for developing NLP applications and LLMs for Sinhala language tasks.

The objective of this study is to evaluate the performance of recent LLMs on several common NLP tasks for the Sinhala language. By systematically testing and fine-tuning these models on Sinhala datasets, we aim to assess their capabilities and limitations when applied to a low-resourced language. Additionally, we explore strategies for adapting and fine-tuning these models to enhance their performance on Sinhala language tasks, such as text generation, summarization, and translation.

Developing and fine-tuning Large Language Models (LLMs) for low-resource languages like Sinhala is crucial for expanding access to digital communication among communities where English proficiency is limited. Additionally, these efforts are essential for preserving cultural and linguistic diversity, enabling tasks such as literature writing, songwriting, and other creative endeavors that rely heavily on native language proficiency and expression.

Through this research, we aim to contribute to the growing body of knowledge on the applicability of LLMs to low-resourced languages and provide insights into the challenges and potential solutions for leveraging these powerful models in resource-constrained settings. The findings of this study will be valuable for researchers, developers, and practitioners working on NLP applications for Sinhala and other low-resourced languages.

## II. Large Language Models

### A. Pre-Trained Multilingual Models

Recent studies have focused on evaluating the performance, consistency, and cross-lingual generalization of pre-trained multilingual language models (LLMs). However, as these models gain widespread application and have downstream societal impact, it is crucial to scrutinize them for fairness across languages, like monolingual models. The paper "Are Pretrained Multilingual Models Equally Fair Across Languages?" [1] investigates this aspect by introducing MozArt, a new multilingual dataset of parallel cloze test examples with demographic annotations. The authors evaluate three multilingual models (mBERT, XLM-R, and mT5) on MozArt and demonstrate that these models exhibit different levels of group disparity across the four target languages, highlighting the importance of evaluating fairness across languages.

Additionally, research has explored addressing the out-of-vocabulary (OOV) problem in multilingual LLMs, which can hinder their performance, especially for low-resource languages. The paper "Improving Pre-Trained Multilingual Models with Vocabulary Expansion" [2] investigates joint mapping and mixture mapping approaches based on the pre-trained multilingual model BERT to tackle the OOV issue. Experimental results across various tasks, including part-of-speech tagging, named entity recognition, and machine translation quality estimation, demonstrate the effectiveness of the mixture mapping approach and the usefulness of bilingual information.

### B. Fine Tuning LLMs

Fine-tuning refers to the process of adapting a pretrained language model's weights to a specific downstream task using task data. The survey [3] discusses several fine-tuning techniques: vanilla fine-tuning directly on the target task, intermediate fine-tuning on a large, related dataset first, multi-task fine-tuning across multiple auxiliary tasks, and parameter-efficient approaches like adapters and pruning that selectively update model components. These methods can improve performance on low-resource target tasks by transferring knowledge from other datasets/tasks and mitigating overfitting.

The paper 'Memory-Efficient Fine-Tuning of Compressed Large Language Models via sub-4-bit Integer Quantization' [4] introduces Parameter-Efficient and Quantization-aware Adaptation (PEQA), a method that combines parameter-efficient fine-tuning and quantization for large language models (LLMs). PEQA fine-tunes only the quantization scales of quantized LLMs while keeping integer weights frozen. This reduces memory usage during training and deployment while allowing efficient task switching. Experiments show PEQA achieves competitive performance to full-precision baselines on various tasks, even with LLMs quantized to sub-4-bit precision, by preserving and enhancing the models' comprehension capabilities.

'Full Parameter Fine-Tuning for Large Language Models with Limited Resources' [5] proposes Low-Memory Optimization (LOMO), a new optimizer that reduces memory usage for full parameter fine-tuning of large language models (LLMs) on limited GPU resources. LOMO fuses gradient computation and parameter updates to minimize gradient tensor sizes. Integrating techniques like gradient normalization allows LOMO to enable full fine-tuning of a 65B LLM on just 8 RTX 3090 GPUs. Experiments on the SuperGLUE benchmark show LOMO can effectively fine-tune massive LLMs while using only 10.8% of the memory required by standard approaches.

### C. Fine Tuning and Prompt Engineering

In a study by Trad and Chehab (2024) [7], the authors compared the effectiveness of prompt engineering and fine-tuning strategies for large language models (LLMs) in the context of phishing URL detection. They explored various prompt-engineering techniques, such as zero-shot, role-playing, and chain-of-thought prompting, and applied them to chat models such as GPT-3.5-turbo and Claude 2. The best prompt-engineering approach achieved an F1-score of 92.74% on a test set of 1,000 samples. They then fine-tuned text generation LLMs such as GPT-2, Bloom, and DistilGPT-2 for sequence classification on phishing detection. Fine-tuning outperformed prompt engineering, with GPT-2 attaining a 97.29% F1-score, surpassing existing state-of-the-art techniques. In realistic scenarios with varying phishing URL ratios, fine-tuned LLMs significantly outperformed prompt-engineered models when phishing URLs were less prevalent than legitimate ones. While prompt engineering facilitates AI system development without training models, fine-tuning LLMs for specific tasks yields superior performance. The authors suggest exploring hybrid approaches combining both strategies, addressing adversarial resilience, real-time optimization, bias mitigation, and incorporating multimodal cues for enhanced phishing detection accuracy.

## III. RECENT LLMs

### A. Claude 3 Sonnet

Claude, an AI assistant developed by Anthropic, excels in versatile conversations and task assistance. The 'Claude 3 Sonnet' model is freely accessible online. Anthropic offers a premium 'Claude 3 Opus' version and recently launched 'Claude 3.5 Sonnet'. [8] This latest model reportedly sets new standards in GPQA (Graduate-Level Google-Proof Q&A), MMLU (Multi-task Language Understanding), and coding benchmarks. [14] However, our research exclusively employed the freely available 'Claude 3 Sonnet' model for analysis and evaluation purposes.

### B. GPT 4o

GPT-4o is Open-AI's newest flagship model, claiming to provide GPT-4(Open AI's commercial offering) level intelligence with improved speed and enhanced capabilities across text, voice, and vision modalities [9]. Open AI also has GPT 3.5 available, which is freely accessible. We explored GPT-4o, which is currently available for free access with certain limitations.

### C. Llama 3

Meta's Llama 3 is a free and open-source large language model available in 8B and 70B parameter versions. It allows developers to customize and deploy the model locally or in the cloud, with performance comparable to proprietary foundation models. Our study utilized the freely available 8B parameter version. Meta is currently developing an even more extensive 400B parameter model, but it was excluded from our study as it is still in the training phase. [10]

### D. Mistral 7B

Mistral AI is a research company developing advanced open-source and commercial language models. They claim to be specialized in creating multilingual AI systems with exceptional capabilities in code generation, mathematical problem-solving, and complex reasoning tasks.[11]We utilized the Mistral 7B model, which is a free and open-source model. Mistral also offers sparse mixture-of-experts models, such as Mixtral 8*7B and Mixtral 8*22B, which are larger and potentially more capable models. However, for the scope of this research, we focused on the more computationally efficient Mistral 7B model.

### E. Falcon 2

Falcon AI is an open-source, transparent 40 billion parameter large language model developed by the UAE's Technology Innovation Institute. It was trained on 1 trillion tokens and provides advanced language capabilities through its innovative architecture [12]. we were unable to successfully set up and access the Falcon model. Efforts were made to set up the base Falcon model on Google Colab, but due to resource constraints and persistent errors during the setup process, we could not establish a functional environment for evaluating Falcon. Additionally, there was a lack of a user-friendly interface or web-based access for Falcon. As a result, Falcon was ultimately excluded from the comparative analysis.

*F. Gemini*

Gemini is Google's new multimodal AI model that can process text, code, audio, images and video. It claims to offer problem-solving prowess, surpassing human experts on benchmarks, while being flexible to run efficiently across devices [13]. The Gemini model encompasses variants such as Gemini Ultra, Pro, Nano, and Flash which are optimized for specific tasks [15]. Although, Gemini lacks multilingual support. Preliminary experiments revealed that none of the Gemini variants could generate outputs in the Sinhala language. Consequently, further experiments involving the Gemini model were deemed unnecessary and excluded from the present study.

Google has also announced Gemma, an open model built from the same research and technology used to create the Gemini models. However, we did not investigate Gemma in this study.

## IV. EVALUATION OF LLMS

*A. Selection of LLMs*

Of the LLMs considered in Section III, we did not test Gemini as it does not support Sinhala. Falcon was also not tested due to operational reasons. Accordingly, we performed our experiments on **Claude 3 Sonnet, GPT 4o, Llama 3 7B** and **Mistral 7B**, all of which were released or updated within the past 12 months. The same prompts were used for each of the LLMs, as described below.

As Claude and GPT are closed models, we only accessed them via the chat interfaces.

As Llama and Mistral provided open models, these were downloaded and run on Google Colab after quantization. The quantized versions would necessarily have poorer performance than the orginals, but we did not compare the quantized and original models. However, other than the fine-tuning in 3) below, we did not do any other modifications or optimizations to the models.

*B. Test Cases*

We conducted experiments to assess the performance of Large Language Models (LLMs) in supporting both Sinhala and English languages. Our evaluation focused on text-based inputs and outputs, comparing the results between the two languages.

Experiments chosen are listed below with the specific reason for choosing them:
1. Translation helped us identify the level of language support.
2. Summarization allowed us to evaluate the model's understanding of each language.
3. Text generation provided insights into the amount of data available for each language.

Additionally, we fine-tuned some models for text summarization using a small, publicly available news dataset.[6] The purpose of fine-tuning is to assess the models' ability to be fine-tuned and to determine whether fine-tuning improved the models' performance. This comprehensive approach allowed us to evaluate various aspects of LLMs' capabilities in handling Sinhala and English languages, providing valuable insights for our research.

A panel of six reviewers, including four professional content writers and translators in Sinhala, evaluated all experimental results to ensure a thorough and credible assessment of the LLMs' performance. All input texts, prompts and results can be found in This HuggingFace Repository.

*1) Translation from Sinhala to English and English to Sinhala*

We evaluated the performance of six available translation systems: i.e., Google, Bing, SiTa[1], Claude 3 Sonnet, GPT 4o, Llama 3 70B and Mistral Large. We included Google, Bing and SiTa to compare the performance of the LLMs with existing translation systems.

Each of these systems was evaluated using their public web interface (i.e., we did not use Colab for the translation experiments). Each system translated (1) three Sinhala texts of approximately 250 words into English, and (2) three English texts of approximately 100 words into Sinhala.

The translations were sent to the panel of reviewers, who assigned a score of 0 = none or nonsense, 1 = very poor to 5 = very good, for each translation.

*2) Text Summarization (untuned):*

We provided three test cases in the Sinhala language of approximately 250 words each. They were on diverse topics: (1) economics, (2) cultural sites and (3) AI in agriculture.

Two experiments were conducted on each of these cases.

  *a)* Direct translation - We gave each model the following prompt:
> Write a 100-word summary of the key points and main ideas in the Sinhala text. The output should be in Sinhala. The summary should not repeat the original text. It should convey the idea of the original text.

The output was obtained in Sinhala Language.

  *b)* We translated the test cases to English and instructed each model to summarize them in English. The summaries were then translated back into Sinhala. The translations were performed using the better of: i. the same model as used for summarization or ii. Google translate.

This allowed each system to utilize its capabilities in English for summarization and allowed us to use a best-performing system for translation.

*3) Text Summarization with Fine Tuning*

The Llama 3 and Mistral models were fine-tuned using a Sinhala news data set which contains approximately 3250 news articles and their summaries [6]. Although this data set is out-of-domain, the fine tuning was used to identify whether these LLMs are amenable to fine tuning using relatively small data sets.

---
[1] SiTa is an internally developed translation system based on mBART.

The other LLMs were not fine-tuned but will be included in future work.

The same prompts and data as in the previous case were applied to each of the two fine-tuned models, and Sinhala summaries obtained.

*4) Text Generation*

Three sets of instructions were provided – to (1) write a travel blog post on Sri Lanka, (2) design a geography assignment on the Amazon Rainforest and (3) write step-by-step guide on how to make a delicious chicken sandwich.

For each of these topics, in addition to direct generation in Sinhala, English prompts were also provided, and the English texts generated were translated into Sinhala.

Each of the experiments was performed on each of the four systems, where possible.

The outputs (summarizations or generations) were sent to the panel of reviewers who assigned each a score on a scale of 0 -5. The average score for each input test case and the overall score for each experiment were calculated.

## V. RESULTS

### A. Translation

Table I shows the average ratings of the panel for each system for translation from Sinhala to English.

S1, S2, and S3 indicate the average score for each of the three Original Sinhala Text phrases used in each experiment. The overall average is calculated using these three scores.

**Table I: TRANSLATION FROM SINHALA TO ENGLISH**

| System | S1 | S2 | S3 | Average |
|---|---|---|---|---|
| Google | 3.5 | 3.3 | 4.0 | 3.6 |
| Bing | 2.6 | 2.8 | 2.6 | 2.7 |
| SiTa | 2.5 | 2.3 | 2.4 | 2.4 |
| Claude | 3.1 | 3.5 | 3.6 | 3.4 |
| GPT4o | 3.8 | 3.4 | 3.5 | 3.5 |
| Llama | 2.5 | 1.5 | 2.5 | 2.2 |
| Mistral | 1.0 | 0.0 | 0.3 | 0.4 |

Table II shows the average ratings for each system for translation from Sinhala to English.

**Table II: TRANSLATION FROM ENGLISH TO SINHALA**

| System | S1 | S2 | S3 | Average |
|---|---|---|---|---|
| Google | 3.3 | 4.2 | 3.7 | 3.7 |
| Bing | 2.5 | 4.0 | 3.7 | 3.4 |
| SiTa | 2.3 | 1.7 | 2.7 | 2.2 |
| Claude | 2.7 | 2.7 | 2.0 | 2.4 |
| GPT4o | 2.3 | 2.3 | 3.0 | 2.6 |
| Llama | 2.8 | 3.5 | 3.0 | 3.1 |
| Mistral | 0.3 | 0.7 | 1.3 | 0.8 |

### B. Text Summarization

Table III shows the average ratings by the review panel for each summarized text for each LLM.

**Table III: TEXT SUMMARIZATION**

| System | Text | Direct | Translated | FineTuned |
|---|---|---|---|---|
| Claude | S1 | 3.8 | 3.6 | |
| | S2 | 3.8 | 4 | |
| | S3 | 4.3 | 3.8 | |
| | Average | 4 | 3.8 | |
| GPT4o | S1 | 3.8 | 3.4 | |
| | S2 | 4 | 4 | |
| | S3 | 4.2 | 5 | |
| | Average | 4 | 4.1 | |
| Llama 3 | S1 | 2 | 1 | 3.5 |
| | S2 | 2.8 | 1 | 2.5 |
| | S3 | 2.9 | 1.5 | 2.5 |
| | Average | 2.6 | 1.2 | 2.8 |
| Mistral | S1 | 0 | 1 | 2.5 |
| | S2 | 0 | 0.8 | 1.5 |
| | S3 | 0 | 1.4 | 2 |
| | Average | 0 | 1.1 | 2 |

### C. Text Generation

Table IV shows the average ratings by the review panel for each generated text for each LLM. P1, P2 and P3 are the prompts used for generating text.

**Table IV: TEXT GENERATION**

| System | Text | Direct | Translated |
|---|---|---|---|
| Claude | P1 | 2.3 | 4.5 |
| | P2 | 3.8 | 4.8 |
| | P3 | 2.5 | 2.3 |
| | Average | 2.9 | 3.9 |
| GPT 4o | P1 | 3.2 | 4.8 |
| | P2 | 4 | 4.8 |
| | P3 | 1.8 | 3 |
| | Average | 3 | 4.2 |
| Llama 3 | P1 | 1.7 | 2.5 |
| | P2 | 2.3 | 2 |
| | P3 | 1.3 | 1 |
| | Average | 3 | 1.8 |
| Mistral | P1 | 0 | 4 |
| | P2 | 0 | 4 |
| | P3 | 0 | 3 |
| | Average | 0 | 3.7 |

These results are discussed in the next section.

## VI. COMPARISON OF MODELS

### A. Translation

For Sinhala-to-English translation, Google, GPT and Claude performed best, though none were able to get an average rating of over 4, showing that they are not usable for production applications. Despite the significant improvements in the performance of machine learning based translation systems in the past few years, more work is needed before they can be used in critical applications in Sinhala.

Bing, SiTa, and Llama managed to translate text – but poorly. Mistal was unable to produce any results.

For English to Sinhala translations, Google was the leader, but missed an average rating of 4. Bing and – surprisingly – Llama, also provided reasonable translations, while GPT, Claude and Sita gave poor results. Mistral did not perform.

Bing and Claude gave incorrect output by omitting the ZWJ (zero-width joiner) character. This character is essential when rendering Sinhala, but some repositories and engines unnecessarily drop this character. The other systems handled this character correctly.

The surprising performance of Llama in the English-to-Sinhala direction – while its Sinhala-to-English performance is poor– requires further investigation. This result indicates that Llama has good text generation capabilities in Sinhala, although officially, Meta does not support Llama in this language.

This experiment indicates that Mistral has not been trained for translation between this pair of languages. However, it may perform better after fine tuning.

*B. Text Summarization*

Both Claude and GPT 4o perform well in direct text summarization. Although GPT is better known, we see that Claude gives similar performance in direct text summarization. Since summarization requires a good 'knowledge' of language, this indicates that Claude has been well-trained in Sinhala. It may also mean that Claude (and other models) use translation (via English or another) internally, when asked to summarize text. Some of the artifacts encountered indicate that the models may do this, although further investigation is needed before a conclusion can be drawn.

Claude tended to make Unicode errors when summarizing.

We see that although both systems make some errors, they may be used for non-mission-critical applications without any change. However, neither is yet quite ready for prime time.

Summarization by translation via English did not result in any significant differences over direct summarization, showing that both these models support a low-resourced language well. However, we suspect that this may be due to the models internally translating the input and output texts to English or an internal representation, due to artifacts we have observed in the direct translation. However, this observation needs further study.

Llama 3 performed poorly in direct summarization, and even worse via translation. However, we show a small improvement by using fine tuning. Therefore, it may be possible to obtain good results from Llama 3 in a specific domain by appropriate fine tuning.

Mistral was not able to perform direct summarization in Sinhala. Its performance via translation was also very poor. However, it showed improvement when fine-tuned, though still far below Claude and GPT 4o.

The above results show that the Llama and Mistral models are fine-tunable and may provide good results after fine-tuning with more data.

*C. Text Generation*

For direct text generation too, Claude and GPT 4o provided the best performance, though significantly below the performance of text summarization and below an acceptable level. Interestingly, both Claude and GPT 4o performed much better at generation via translation, giving good results. This may mean that these models have not been trained much to work with Sinhala prompts.

This also provides us with a promising area of inquiry – how to improve LLM performance using pre- and post-processing techniques such as translation and retrieval-augmented generation.

Llama 3 yielded poor results, both directly and via translation Mistral did not support direct text generation in Sinhala but gave reasonable results via translation.

VII. SUMMARY

This study shows that two of the major LLMs – Claude and GPT – support Sinhala well. This experiment was conducted on their small (~7B) models, but their large models may perform significantly better. Unfortunately, as both these are closed models, we cannot fine tune them. However, we plan to investigate how techniques such as prompt engineering and RAG may enhance their performance.

Although we did not study other low-resourced languages, we surmise that these two models also support other languages with a similar level of resources.

Although Llama 3 was announced with much fanfare, it – and Mistral – has little support for Sinhala, except in translation to that language.

As this is a preliminary investigation, we plan to continue to study these, and other LLMs, in detail to identify how well they work with low-resourced languages.

We have shown that the major LLMs introduced in 2024 have much better support for Sinhala (and possibly other low-resourced languages) than the previous crop. We also see that while the out-of-the-box performance of the open models is poor, there is scope for improvement by using fine tuning and other techniques.

REFERENCES

[1] L. Cabello Piqueras and A. Søgaard, "Are Pretrained Multilingual Models Equally Fair across Languages?" in *Proceedings of the 29th International Conference on Computational Linguistics*, N. Calzolari, C.-R. Huang, H. Kim, J. Pustejovsky, L. Wanner, K.-S. Choi, P.-M. Ryu, H.-H. Chen, L. Donatelli, H. Ji, S. Kurohashi, P. Paggio, N. Xue, S. Kim, Y. Hahm, Z. He, T. K. Lee, E. Santus, F. Bond, and S.-H. Na, Eds., Gyeongju, Republic of Korea: International Committee on Computational Linguistics, Oct. 2022, pp. 3597–3605. Available: https://aclanthology.org/2022.coling-1.318. [Accessed: May. 12, 2024]

[2] H. Wang, D. Yu, K. Sun, J. Chen, and D. Yu, *Improving Pre-Trained Multilingual Models with Vocabulary Expansion*. 2019. Available: https://www.researchgate.net/publication/336132647_Improving_Pre-Trained_Multilingual_Models_with_Vocabulary_Expansion

[3] K. S. Kalyan, A. Rajasekharan, and S. Sangeetha, "AMMU: A survey of transformer-based biomedical pretrained language models," *Journal of Biomedical Informatics*, vol. 126, p. 103982, Feb. 2022, doi: 10.1016/j.jbi.2021.103982. Available: https://www.sciencedirect.com/science/article/pii/S1532046421003117. [Accessed: May. 12, 2024]

[4] J. Kim *et al.*, "Memory-efficient fine-tuning of compressed large language models via sub-4-bit integer quantization," in *Proceedings of the 37th International Conference on Neural Information Processing Systems*, in NIPS '23. Red Hook, NY, USA: Curran Associates Inc., May 2024, pp. 36187–36207. Available: https://dl.acm.org/doi/10.5555/3666122.3667691

[5] K. Lv, Y. Yang, T. Liu, Q. Gao, Q. Guo, and X. Qiu, *Full Parameter Fine-tuning for Large Language Models with Limited Resources*. 2023. Available: https://www.researchgate.net/publication/371684938_Full_Parameter_Fine-tuning_for_Large_Language_Models_with_Limited_Resources

[6] Ziyard Hamza, "Sinhala News articles and summaries from BBC", HaggingFace, Available: https://huggingface.co/datasets/Hamza-Ziyard/BBC-Sinhala [Accessed: May, 2024]

[7] F. Trad and A. Chehab, "Prompt Engineering or Fine-Tuning? A Case Study on Phishing Detection with Large Language Models," *Machine Learning and Knowledge Extraction*, vol. 6, no. 1, pp. 367–384, Mar. 2024, doi: 10.3390/make6010018. Available: https://www.mdpi.com/2504-4990/6/1/18. [Accessed: May. 12, 2024]

[8] ["Models," *Anthropic*. Available: https://docs.anthropic.com/en/docs/about-claude/models. [Accessed: May. 10, 2024]

[9] "Hello GPT-4o." Available: https://openai.com/index/hello-gpt-4o/. [Accessed: May. 10, 2024]

[10] "Getting started with Llama | Documentation." Available: https://llama.meta.com/docs/get-started/. [Accessed: May. 10, 2024]

[11] "Bienvenue to Mistral AI Documentation | Mistral AI Large Language Models." Available: https://docs.mistral.ai/. [Accessed: May. 10, 2024]

[12] "Falcon LLM." Available: https://falconllm.tii.ae/. [Accessed: May. 10, 2024]

[13] "Gemini API Developer Docs and API Reference," *Google for Developers*. Available: https://ai.google.dev/gemini-api/docs. [Accessed: May. 10, 2024]

[14] "Introducing Claude 3.5 Sonnet." Accessed: Jul. 29, 2024. [Online]. Available: https://www.anthropic.com/news/claude-3-5-sonnet

[15] "Gemini - Google DeepMind." Available: https://deepmind.google/technologies/gemini/. [Accessed: Jul. 06, 2024]